\documentclass[conference,a4paper]{ieeetran}

\usepackage{times}
\usepackage{balance}
\usepackage{setspace}
\usepackage{algorithm}
\usepackage{algorithmic}
\usepackage{relsize}
\usepackage{boxedminipage} 
\usepackage{graphicx}
\usepackage{verbatim} 
\usepackage{amsmath}
\usepackage{times}
\usepackage{gensymb}
\usepackage{dsfont}
\usepackage{multirow}
\usepackage{float}
\usepackage{amsfonts}
\usepackage{epstopdf}
\usepackage{mathrsfs}
\usepackage[T1]{fontenc}
\usepackage{url}
\usepackage{amssymb}
\usepackage[tight,footnotesize]{subfigure}
\usepackage[caption=false,font=footnotesize]{subfig}
\usepackage{epsfig}
\usepackage{csquotes}
\usepackage{array}
\usepackage[scientific-notation=false]{siunitx}
\usepackage{booktabs}

\usepackage[table,x11names]{xcolor}

\newcolumntype{L}[1]{>{\raggedright\let\newline\\\arraybackslash\hspace{0pt}}m{#1}}
\newcolumntype{C}[1]{>{\centering\let\newline\\\arraybackslash\hspace{0pt}}m{#1}}
\newcolumntype{R}[1]{>{\raggedleft\let\newline\\\arraybackslash\hspace{0pt}}m{#1}}

\title{Dynamic Ensemble Selection VS K-NN: why and when Dynamic Selection obtains higher classification performance?}

\author{\authorblockN{Rafael M. O. Cruz\authorrefmark{1}, Hiba H. Zakane\authorrefmark{1}, Robert Sabourin\authorrefmark{1} and George D. C. Cavalcanti\authorrefmark{2} }
	\authorblockA{\authorrefmark{1}\'{E}cole de Technologie Sup\'{e}rieure - Universit\'{e} du Qu\'{e}bec, Canada\\
		e-mail: rafaelmenelau@gmail.com, zakanehiba@gmail.com,  Robert.Sabourin@etsmtl.ca}
	\authorblockA{\authorrefmark{2}Centro de Inform\'{a}tica - Universidade Federal de Pernambuco, Brazil\\
		e-mail: gdcc@cin.ufpe.br}}

\IEEEoverridecommandlockouts
 
\pubid{\makebox[\columnwidth]{978-1-5386-1842-4/17/\$31.00~\copyright{} 2017 IEEE \hfill}\hspace{\columnsep}\makebox[\columnwidth]{ }}

\begin{document}
	\maketitle
	\begin{abstract}
		Multiple classifier systems focus on the combination of classifiers to obtain better performance than a single robust one. These systems unfold three major phases: pool generation, selection and integration. One of the most promising MCS approaches is Dynamic Selection (DS), which relies on finding the most competent classifier or ensemble of classifiers to predict each test sample. The majority of the DS techniques are based on the K-Nearest Neighbors (K-NN) definition, and the quality of the neighborhood has a huge impact on the performance of DS methods. In this paper, we perform an analysis comparing the classification results of DS techniques and the K-NN classifier under different conditions. Experiments are performed on 18 state-of-the-art DS techniques over 30 classification datasets and results show that DS methods present a significant boost in classification accuracy even though they use the same neighborhood as the K-NN. The reasons behind the outperformance of DS techniques over the K-NN classifier reside in the fact that DS techniques can deal with samples with a high degree of instance hardness (samples that are located close to the decision border) as opposed to the K-NN. In this paper, not only we explain why DS techniques achieve higher classification performance than the K-NN but also when DS should be used. 
	\end{abstract}
	
	
	\section{Introduction}
	
	One of the most promising MCS approaches is Dynamic Selection (DS), in which the base classifiers\footnote{The term base classifier refers to a single classifier belonging to an ensemble or a pool of classifiers.} are selected on the fly, according to each new sample to be classified. DS has become an active research topic in the multiple classifier systems literature in the past years. This is due to the fact that more and more works are reporting the superior performance of such techniques over traditional combination methods, such as Majority Voting and Boosting~\cite{CRUZ2018195,Alceu2014,Woloszynski,knora,CruzPR}. DS techniques work by estimating the competence level of each classifier from a pool of classifiers. Only the most competent, or an ensemble containing the most competent classifiers, is selected to predict the label of a specific test sample. The rationale for such techniques is that not every classifier in the pool is an expert in classifying all unknown samples; rather, each base classifier is an expert in a different local region of the feature space~\cite{zhu}. 
	
	In dynamic selection, the key is how to select the most competent classifiers for any given query sample. The competence of the classifiers is estimated based on a local region of the feature space where the query sample is located, called region of competence. This region is usually defined by applying the K-Nearest Neighbors technique to find the neighborhood of this query sample. Then, the competence level of the base classifiers is estimated, considering only the samples belonging to the region of competence according to any selection criteria; these include the accuracy of the base classifiers in this local region~\cite{lca,Smits_2002,SoaresSCS06} or ranking~\cite{classrank} and probabilistic models~\cite{Woloszynski,WoloszynskiKPS12}. The classifier(s) that attained a certain competence level is(are) selected.
	
	Several works pointed out that the performance of DS techniques is very sensitive to the definition of the region of competence~\cite{ijcnn2011,reportarXiv,oliveira2002automatic}. If there is a noise in the defined neighborhood of the query sample, the DS systems are more likely to fail. Moreover, the use of different K-NN approaches for the definition of the regions of competences can significantly change the performance of DS methods~\cite{cruz2016prototype}.
	
	As the competence of the base classifiers are heavily dependent on the K-Nearest Neighbors for the definition of the local regions, one question arises: Why do we use dynamic selection instead of simply applying the K-NN classifier? Moreover, in which scenario the use of DS brings benefits over the K-NN? To the best of our knowledge, there is no comparison between both classification approaches in the DS literature. Hence, the objective of this paper is to perform an analysis comparing the classification results of DS techniques and the K-NN classifier. In particular, the following points are investigated:
	
	\begin{enumerate}
		\item Do DS techniques achieve higher classification performance than the K-NN?
		\item Why does DS present better classification accuracy than K-NN even though the same neighborhood is considered for both techniques?
		\item When should DS be used for classification instead of K-NN?
	\end{enumerate}
	
	Experiments are carried out using 18 state-of-the-art DS technique over 30 classification datasets. We demonstrate that not only DS techniques achieves significantly better results, but we also demonstrate in which scenarios DS techniques can improve the generalization performance over the K-NN classifier.
	
	This paper is organized as follows: Section~\ref{sec:ds} presents the related works on dynamic selection. Section~\ref{sec:experiments} addresses the experiments conducted on state-of-the-art DS techniques. Conclusion and future works are presented in the last section.
	
	\section{Dynamic selection}
	\label{sec:ds}
	
Dynamic selection techniques consist, based on a pool of classifiers $C$, in finding a single classifier $c_{i}$, or an ensemble of classifiers $C'\subset C$, that has (or have) the most competent classifiers to predict the label for a specific test sample, $\mathbf{x}_{q}$. The most important component of DES techniques is how the competence level of the base classifier is measured, given a specific test sample $\mathbf{x}_{q}$. This is a different concept from static selection methods~\cite{classmaj,SantosSM09}, in which the Ensemble of Classifiers (EoC), $C^{'}$, is selected during the training phase, according to a selection criterion estimated in the validation dataset, and is used to predict the label of all test samples in the generalization phase.

In dynamic selection, the classification of a new query sample normally involves three phases: 

\begin{enumerate}
	\item The definition of the region of competence; that is, how to define the local region surrounding the query, $\mathbf{x}_{q}$, in which the competence level of the base classifiers is estimated
	
	\item The selection criteria used to estimate the competence level of the base classifiers, e.g., Accuracy, Probabilistic, and Ranking
	
	\item The selection mechanism that chooses a single classifier (DCS) or an ensemble of classifiers (DES) based on their estimated competence level
\end{enumerate}

The most common method to define the regions of competence is by using the K-NN technique, to get the neighborhood of the test sample~\cite{lca,knora,Smits_2002,giacinto1999methods,DidaciGRM05,mcb,Brun2016,CruzPR,WoloszynskiKPS12,SoaresSCS06,classrank,Oliveira2017}. The set with the K-Nearest Neighbors of a given test sample $\mathbf{x}_{q}$ is called region of competence, and is denoted by $\theta_{q} = \left \{ \mathbf{x}_{1}, \ldots, \mathbf{x}_{K} \right \}$. Many works pointed out that the definition of this region of competence is of fundamental importance to DS methods, as the performance of all DS techniques is very sensitive to the distribution of this region~\cite{cruz2016prototype,cruzijcnn2017}. The samples belonging to $\theta_{q}$ are used to estimate the competence of the base classifiers, for the classification of $\mathbf{x}_{q}$, based on various criteria, such as the overall accuracy of the base classifier in this region~\cite{lca}, ranking~\cite{classrank}, ambiguity~\cite{docs}, oracle~\cite{knora} and probabilistic models~\cite{WoloszynskiKPS12}. In any case, a set of labeled samples, which can be either the training or validation set, is required for the definition of the local regions. This set is called the dynamic selection dataset (DSEL)~\cite{paulo2}. 

After the competence level of the base classifiers are estimated, the most competent one or an ensemble containing the most competent classifiers, to predict the label of $\mathbf{x}_{q}$ is(are) selected. For instance the Overall-Local-Accuracy (OLA)~\cite{lca} and Multiple Classifier Behavior (MCB)~\cite{mcb} techniques select only the classifier that achieved the highest competence level in the neighborhood, while the K-Nearest Oracles techniques (KNORA)~\cite{knora} and the Dynamic Ensemble Selection-Performance (DES-P)~\cite{WoloszynskiKPS12}, and META-DES~\cite{CruzPR} select an EoC containing the most competent classifiers.

As the neighborhood of the query sample is not used directly to predict its label, but rather to estimate the competence level of the base classifiers. This brings benefits when dealing with samples located in an indecision region, i.e., which are located in areas surrounding classes boundaries~\cite{garcia2015dealing}. When the query is located in such a region, the majority of its K-Nearest Neighbors may belong to a different class, which can lead to bad predictions. Moreover, samples located in indecision regions are often misclassified by other pattern recognition techniques since they are usually associated with a high degree of instance hardness~\cite{SmithMG14}.

\begin{figure}[!ht]
	
	\begin{center}  	 
		\includegraphics[clip=,  width=0.50\textwidth]{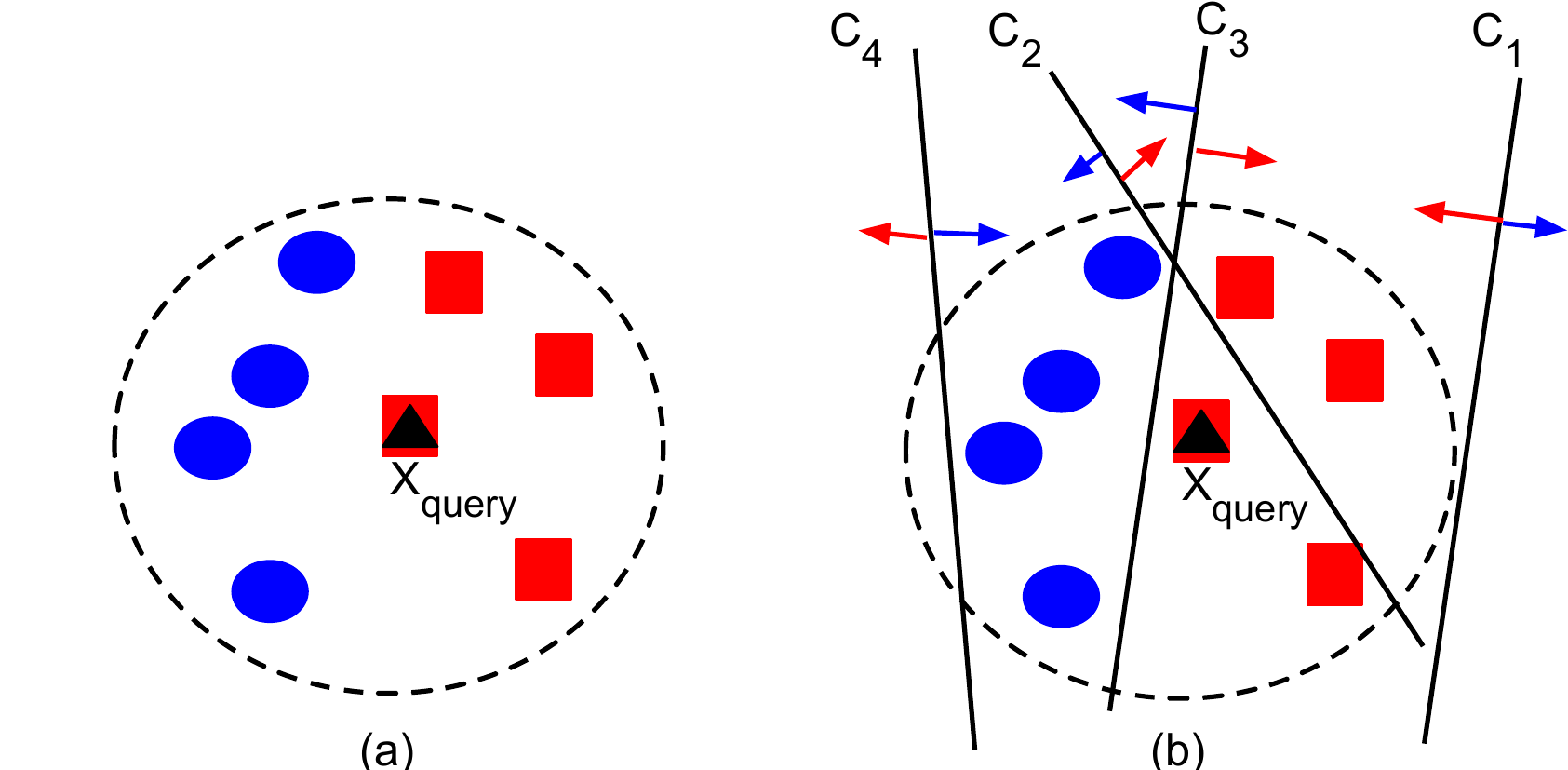}
	\end{center}
	\caption{Example of a query sample located in an indecision region. (a) The estimated region of competence with K = 7. (b) Decision border of different base classifiers with the arrows pointing to the regions of both classes. As the majority of its neighbors belong to a different class, the K-NN classifier would make the wrong prediction. However, DS techniques can still make the right decision if the DS method selects the base classifier that is competent locally ($c_{3}$).}
	\label{fig:regioncompetence}
\end{figure}

However, DS techniques can still predict the correct label for such samples as long as there exists at least one base classifier that is competent locally. In other words, a classifier that can correctly classify samples belonging to different classes in the indecision regions. For example, Figure~\ref{fig:regioncompetence} shows an example of an indecision region. The query sample $\textbf{x}_{query}$, belongs to the class 1 (red square). Since the majority of its neighbors comes from the class 2 (blue circle), a K-NN classifier, considering this whole neighborhood, would misclassify the query sample. 

Using dynamic selection it is possible to predict the correct label of such sample as long as there are base classifiers that cross this indecision region. For instance, consider the system consisting of four base classifiers as shown in Figure~\ref{fig:regioncompetence} (b). If we apply the Overall-Local-Accuracy (OLA) technique~\cite{lca}, the classifier $c_{3}$ would be selected, since it obtained a 100\% accuracy for the local region. The other base classifier that predicted the correct label is $c_{1}$ yet, it does not cross the region of competence, knowing that it achieves a level of competence of 0.33 which is lower than classifiers $c_{2}$ and $c_{4}$ with 0.85 and 0.57 respectively.
Consequently, using dynamic selection it is possible to give the correct prediction for this sample as long as there is at least one base classifier that obtains a high competence level in the local region.

Thus, our hypothesis is that DS techniques outperform the K-NN classifier since it can better deal with samples that are located in indecision regions. This hypothesis is evaluated in the next section.

	\section{Experiments}
	\label{sec:experiments}
		
	\begin{table}[htbp]
		\centering
		\caption{Summary of the 30 datasets used in the experiments [Adapted from~\cite{CruzPR}].}
		\label{table:datasets} 
		\resizebox{.5\textwidth}{!}{
			\begin{tabular}{l r r r  l}
				\hline
				\textbf{Database} & \textbf{ No. of Instances} & \textbf{Dimensionality} & \textbf{No. of Classes}   & \textbf{Source} \\
				\hline
				
				\textbf{Adult} & 48842 & 14 & 2 & UCI  \\        
				
				\textbf{Banana}  & 1000 & 2 &	2 &  PRTOOLS  \\
				
				\textbf{Blood transfusion} & 748 & 4 &	2 &  UCI  \\
				
				\textbf{Breast (WDBC)} & 568 & 30 & 2 &  UCI \\
				
				\textbf{Cardiotocography (CTG)} & 2126 & 21 & 3 &  UCI \\    
				
				\textbf{Ecoli} & 336 & 7 & 8 &  UCI  \\    
				
				\textbf{Steel Plate Faults} & 1941 & 27 & 7 &  UCI \\   
				
				\textbf{Glass} & 214 & 9 & 6  &  UCI  \\                       
				
				\textbf{German credit} & 1000 & 20 &2  &  STATLOG \\
				
				\textbf{Haberman's Survival} & 306 & 3 & 2 & UCI \\
				
				\textbf{Heart} & 270 & 13  & 2  &  STATLOG \\
				
				\textbf{ILPD} & 583 & 10 & 2  &  UCI \\                       
				
				\textbf{Ionosphere} & 315 &	34 & 2 &  UCI  \\
				
				\textbf{Laryngeal1} & 213 & 16 & 2 &  LKC \\        
				
				\textbf{Laryngeal3} & 353 & 16 & 3 &  LKC \\    
				
				\textbf{Lithuanian}  & 1000 & 2 & 2 &  PRTOOLS \\
				
				\textbf{Liver Disorders} & 345 & 6 & 2 & UCI  \\
				
				\textbf{MAGIC Gamma Telescope}  & 19020 & 10 & 2 &  KEEL \\
				
				\textbf{Mammographic}  & 961 & 5 & 2 &  KEEL \\
				
				\textbf{Monk2}  & 4322 & 6 & 2 &  KEEL  \\
				
				\textbf{Phoneme} & 5404 & 6 & 2 &  ELENA  \\   
				
				\textbf{Pima} & 768 & 8 & 2 & UCI  \\
				
				\textbf{Satimage} & 6435 & 19 & 7 & STATLOG \\    
				
				\textbf{Sonar} & 208 &	60 & 2 &  UCI \\
				
				\textbf{Thyroid} &  215 & 5 & 3 &  LKC \\
				
				\textbf{Vehicle} & 846 & 18 & 4 &  STATLOG \\
				
				\textbf{Vertebral Column} & 310 & 6 & 2 &  UCI \\          
				
				\textbf{WDG V1} & 5000 & 21 & 3 &  UCI \\    
				
				\textbf{Weaning} & 302 & 17 & 2 &  LKC \\
				
				\textbf{Wine} & 178 & 13 & 3 &  UCI \\
				\hline
				
			\end{tabular}
		}
	\end{table} 
	
	The comparative study was performed using a test bed composed of 30 classification problems proposed in~\cite{CruzPR}. The key features of the datasets are presented in Table~\ref{table:datasets}.
	For each dataset, the experiments were carried out using 20 replications. For each replication, the datasets were randomly divided on the basis of 25\% for training, $\mathcal{T}$, 50\% for the dynamic selection dataset, $DSEL$, and 25\% for the generalization set, $\mathcal{G}$. The divisions were performed while maintaining the prior probabilities of each class. For the K-NN classifier, $DSEL$ was merged with the training data. As a result, all methods were trained using the same amount of data available, while the distribution of the test set remained the same. The pool of classifiers $C$ was composed of 100 Perceptrons generated using the Bagging technique. The same pool of classifiers was used for all DS techniques. Moreover, the size of the region of competence (neighborhood size) $K$ was equally set at 7 for all techniques since it presented the best classification performance according to~\cite{knora,CruzPR}. 
	
	The analysis is conducted using 18 state-of-the-art DS techniques, eight DCS and ten DES techniques. For DCS, the following techniques were evaluated: Local Class Accuracy (LCA)~\cite{lca}, Overall Local Accuracy (OLA)~\cite{lca}, Modified Local Accuracy (MLA)~\cite{Smits_2002}, Modified Classifier Ranking (RANK)~\cite{classrank,lca}, Multiple Classifier Behavior (MCB)~\cite{mcb}, A Priori~\cite{giacinto1999methods,DidaciGRM05}, A Posteriori~\cite{giacinto1999methods,DidaciGRM05} and the Dynamic Selection on Complexity (DSOC). For dynamic ensemble selection, the following techniques were considered: K-Nearest Oracles Eliminate (KNORA-E)~\cite{knora}, K-Nearest Oracles Union (KNORA-U)~\cite{paulo2}, Randomized Reference Classifier (DES-RRC)~\cite{WoloszynskiK10}, K-Nearest Output Profiles (KNOP)~\cite{paulo2,logid}, Dynamic Ensemble Selection Performance (DES-P)~\cite{WoloszynskiKPS12}, Dynamic Ensemble Selection Kullback-Leibler (DES-KL)~\cite{WoloszynskiKPS12}, DES Clustering~\cite{SoaresSCS06}, DES-KNN~\cite{SoaresSCS06}, Meta Learning for Dynamic Selection (META-DES)~\cite{CruzPR} and META-DES.Oracle~\cite{cruz2017meta}. 
	
	Pseudo-code for the implementation of each method is given in~\cite{Alceu2014,CRUZ2018195}. It is important to point out that 15 out of the 18 DS techniques use the K-NN to define the region of competence, the only exceptions being the DES-RRC, DES-KL and the DES-KMEANS. However, they still use local information in order to estimate the competence level of the base classifiers.
	
	\subsection{Comparison DS vs K-NN}
	
	The first analysis conducted in this paper is a comparison between the accuracy obtained by DS techniques and the K-NN classifier.	The objective of this analysis is to know whether the use of DS leads to a significant improvement in classification accuracy. For the K-NN classifier, we consider a $K = 7$ (i.e., the same neighborhood size used by the DS techniques) as well as the $K = 1$ which is used as a baseline comparison.
		
	Table~\ref{table:rankMeanDESKNN} shows the average ranking and mean accuracy of each technique considering the 30 classification problems studied. The average ranks were obtained using the Friedman test~\cite{Friedman1937Use} as follows: For each dataset, the method that achieved the best performance received rank 1, the second best rank 2, and so forth. In case of a tie, i.e., two methods presented the same classification accuracy for the dataset, their average ranks were summed and divided by two. The average rank was then obtained, considering all datasets. The best performing algorithm, considering the 30 classification datasets, was the one presenting the lowest average rank.
	
	All DS techniques presented a better ranking and average accuracy when compared to the 1-NN, and only the MLA technique presented a lower classification accuracy and lower rank than the K-NN using the same neighborhood size (K=7). This is an interesting finding, since the majority of the DS techniques in this study (14 methods) use the K-NN method in the process of estimating the local competence of the base classifiers. 
	
	\begin{table}[h!] 
		\centering 
		\caption{Overall results} 
		\label{table:rankMeanDESKNN}  
		\resizebox{0.48\textwidth}{!}{  
			\begin{tabular}{|c | c || c| c|}  
				\hline  
				
				Algorithm & Avg. Rank & Algorithm & Avg. Accuracy \\ \hline  
				
				META-DES.O &  4.07(3.67) & META-DES.O & 83.92(9.13) \\ 
				META-DES &  4.40(3.23) & META-DES & 83.24(8.94) \\ 
				DES-RRC &  6.40(5.30) & DES-P & 82.26(9.26) \\ 
				KNORA-U &  7.33(4.65) & DES-RRC & 82.11(8.76) \\ 
				DES-P &  7.57(4.06) & KNORA-U & 81.69(9.82) \\ 
				DES-KL &  8.20(5.43) & DES-KL & 81.52(8.77) \\ 
				KNOP &  10.27(4.19) & KNOP & 80.81(8.92) \\ 
				KNORA-E &  10.40(4.21) & KNORA-E & 80.36(10.75) \\ 
				LCA &  10.80(4.91) & OLA & 79.87(10.67) \\ 
				OLA &  11.07(5.23) & DCS Rank & 79.69(10.38) \\ 
				DSOC &  11.63(6.17) & DSOC & 79.68(9.44) \\ 
				MCB &  11.93(5.39) & LCA & 79.57(9.84) \\ 
				DES-KNN &  12.00(4.72) & MCB & 79.56(9.70) \\ 
				A Posteriori &  12.17(5.68) & DES-KNN & 79.29(10.23) \\ 
				DCS Rank &  12.53(4.53) & A Priori & 78.57(11.18) \\ 
				7NN &  12.97(6.32) & DES-KMEANS & 78.49(10.40) \\ 
				DES-KMEANS &  13.57(4.26) & A Posteriori & 78.14(11.53) \\ 
				MLA &  13.63(5.12) & 7NN & 77.42(13.06) \\ 
				A Priori &  13.77(4.67) & MLA & 77.34(9.78) \\ 
				1NN &  15.30(5.95) & 1NN & 76.64(11.98) \\ 
				\hline  
				
			\end{tabular}} 
		\end{table} 
		
		Furthermore, a pairwise analysis was conducted based on the Sign test~\cite{Demsar:2006}, computed on the number of wins, ties and losses obtained by each DS, compared to the 7-NN (i.e., same neighborhood size). The null hypothesis, $H_{0}$, meant that both techniques obtained statistically equivalent results. A rejection in $H_{0}$ meant that the classification performance obtained by a corresponding DS technique was significantly better at a predefined significance level $\alpha$. In this case, the null hypothesis, $H_{0}$, is rejected when the number of wins is greater than or equal to a critical value, denoted by $n_{c}$. The critical value is computed using Equation~\ref{eq:criticalvalue}
		
		\begin{equation}
		\label{eq:criticalvalue}
			n_{c} = \frac{n_{exp}}{2} + z_{\alpha}\frac{\sqrt{n_{exp}}}{2}
		\end{equation}

		\begin{figure}[!ht]
			\begin{center}  	 
				\includegraphics[clip=,  width=0.5\textwidth]{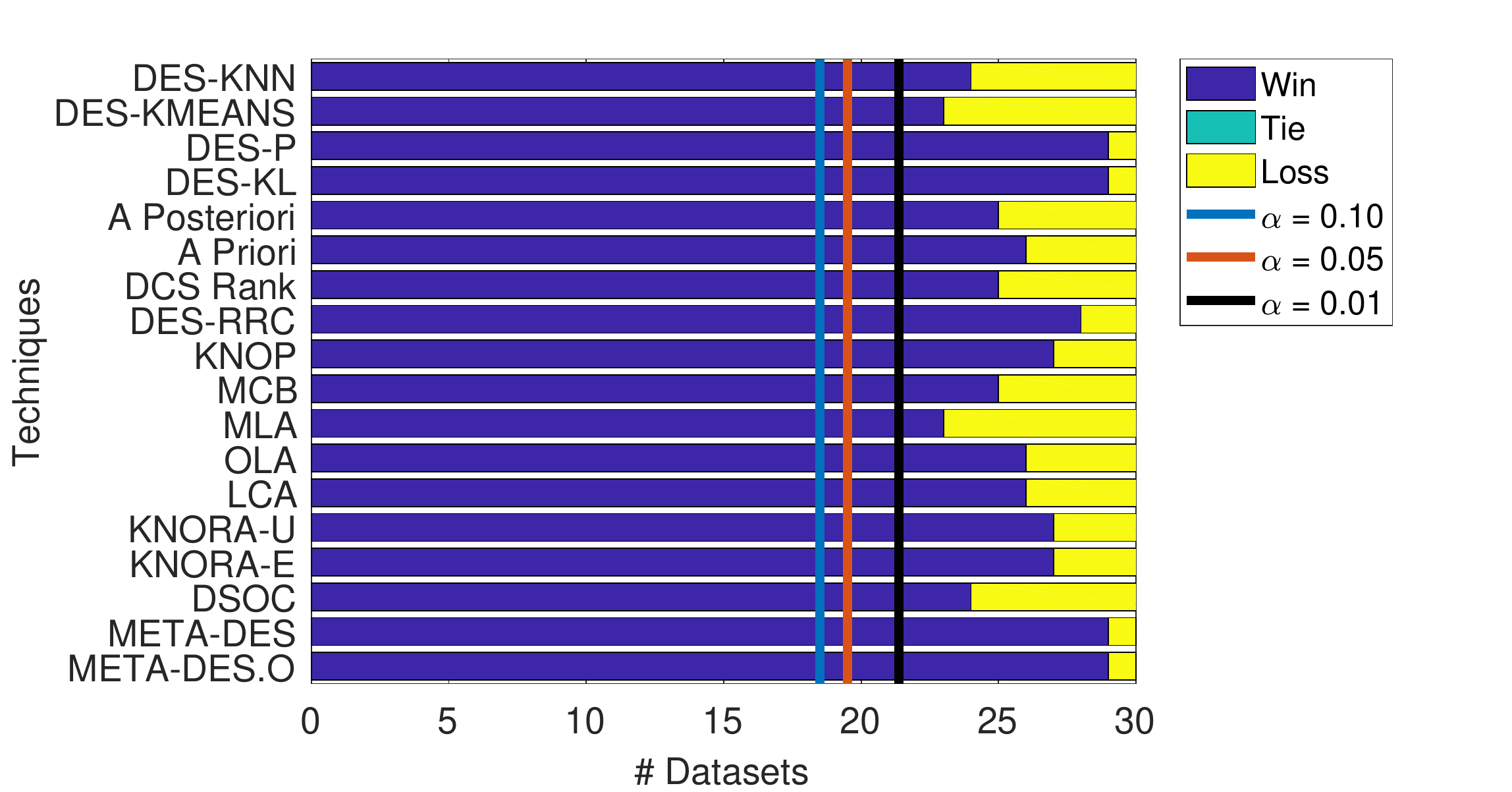}
			\end{center}
			\caption{Pairwise comparison between the results achieved using the different DS techniques and the 1-NN. The analysis is based on wins, ties and losses. The vertical lines illustrate the critical values considering a confidence level $\alpha = \{0.10, 0.05, 0.01 \}$.}
			\label{fig:wintielo1NN}
		\end{figure}
		
		\begin{figure}[!ht] 	
			\begin{center}  	 
				\includegraphics[clip=,  width=0.5\textwidth]{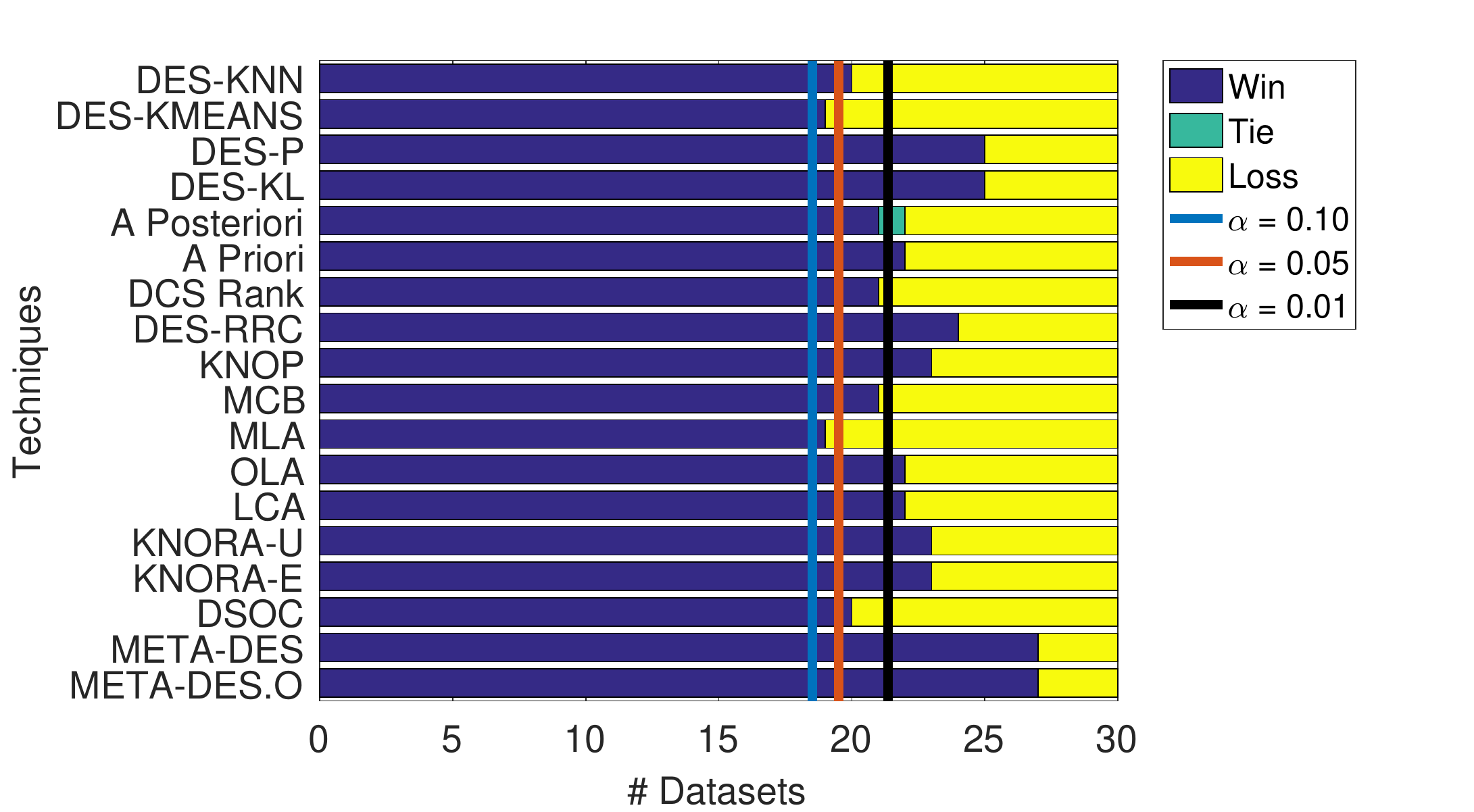}
			\end{center}
			\caption{Pairwise comparison between the results achieved using the different DS techniques and the 7-NN. The analysis is based on wins, ties and losses. The vertical lines illustrate the critical values considering a confidence level $\alpha = \{0.10, 0.05, 0.01 \}$.}
			\label{fig:wintieloKNN}
		\end{figure}   
		
		\noindent where $n_{exp}$ is the total number of experiments. We ran the test considering three levels of significance: $\alpha = \{0.10, 0.05, 0.01 \}$. Figures~\ref{fig:wintielo1NN} and~\ref{fig:wintieloKNN} show the results of the Sign test comparing the performance of DS techniques and the 1-NN and 7-NN respectively. The different bars represent the critical values for each significance level.
		
		Compared to the 1-NN, we can see that all DS methods presented a significant number of wins even when the level of significance is reduced to $\alpha = 0.01$. Compared to the 7-NN (i.e., the same neighborhood size as the DS techniques)  we can see that at a 0.1 significance level, all DS techniques obtained a significant number of wins. Using an $\alpha = 0.05$, only two DS methods (DS-KMEANS and MLA) did not present a significant number of wins. Moreover, even restricting the test to a significance level of 0.01, we could see that the majority of the DS techniques obtained a significant number of wins. Therefore DS methods present a significant boost in classification accuracy even though they use the same neighborhood as the K-NN.

		\subsection{Instance hardness analysis}
		
		Instance hardness (IH) measure provides a framework for identifying which instances are hard to classify and also, understand why they are hard to classify~\cite{SmithMG14}. The objective of this experiment is to analyze the performance of DS techniques and the K-NN classifier for dealing with samples with different degrees of instance hardness. Thus, we want to test our hypothesis that DS techniques can better handle samples that are located in indecision regions, and are associated with a higher degree of instance hardness.
				
		The kDisagreeing Neighbors (kDN) is considered, since it presented the highest correlation with the probability that a given instance is misclassified by different classification methods according to~\cite{SmithMG14}. The
		kDN measure is the percentage of instances in an instance's neighborhood that
		do not share the same label as itself. Equation~\ref{eq:kdn} shows the kDN measure. 
		
		\begin{equation}
		\label{eq:kdn}
		kDN(\mathbf{x}_{q}) = \frac{\mid \mathbf{x}_{k} : \mathbf{x}_{k} \in KNN(\mathbf{x}_{q}) \wedge t(\mathbf{x}_{k}) \neq t(\mathbf{x}_{q})\mid}{K}
		\end{equation}
			
		\noindent where $KNN(\mathbf{x}_{q})$ is the set of $K$ nearest neighbors of $\mathbf{x}_{q}$, and $\mathbf{x}_{k}$ represents an instance in this neighborhood. $t(\mathbf{x}_{q})$ and $t(\mathbf{x}_{k})$ represents the target class of the instances $\mathbf{x}_{q}$ and $\mathbf{x}_{k}$ respectively. 
		
		In this work, we considered a neighborhood size $K = 7$ for the estimation of the kDN, which is the same neighborhood sized used for the DS techniques as well as the K-NN classifier.
		
		We rank the testing instances of all datasets according to their level of IH. Then, the samples were divided into 8 groups (given that $K = 7$) with the possible configurations of IH (starting from IH = 0, when the whole neighborhood agrees with the class of the test sample, up to IH = 1, when the whole neighborhood disagrees with the label of the test sample).
		
\begin{figure}[!ht]
			
			\begin{center}  	 
				\includegraphics[clip=,  width=0.50\textwidth]{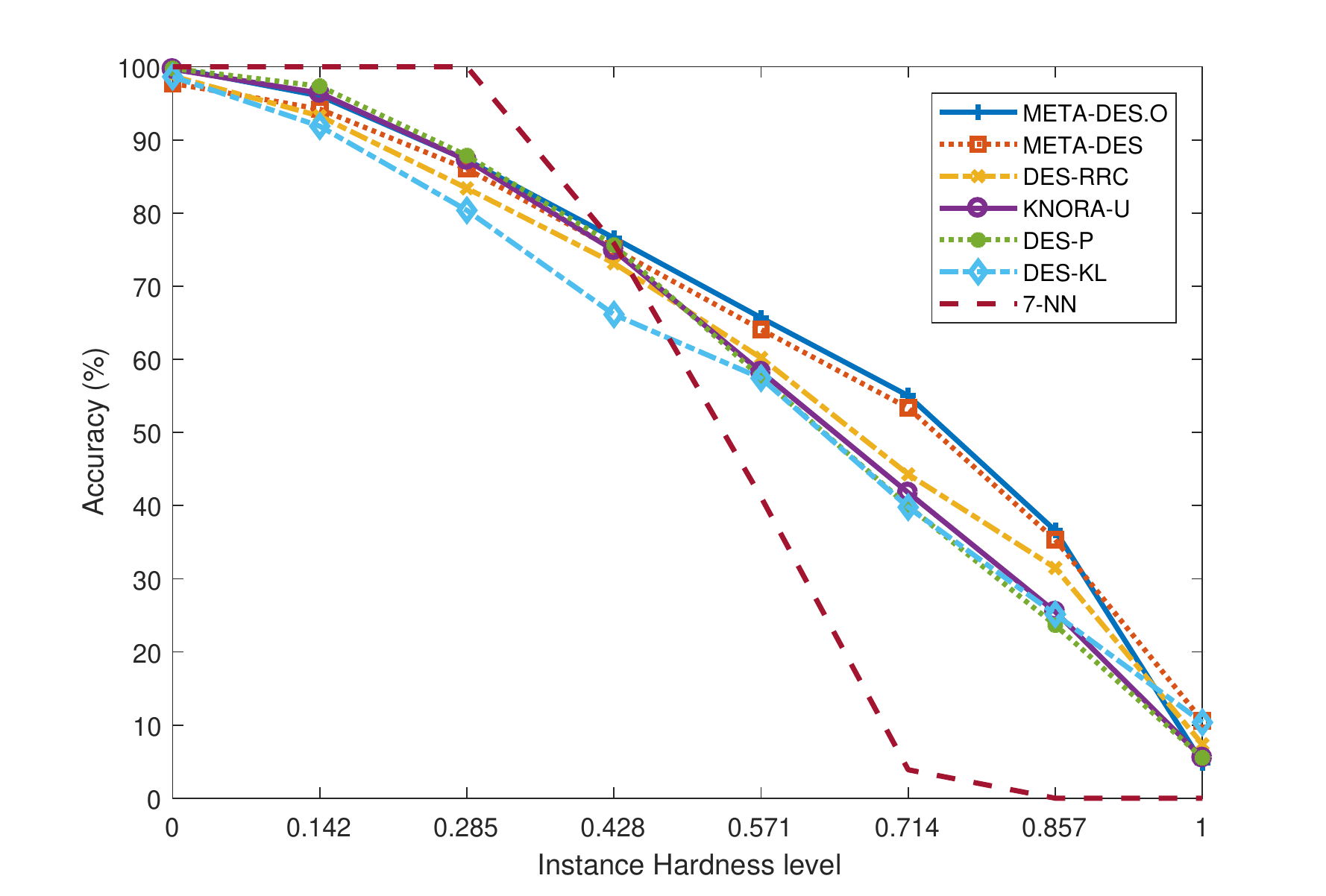}
			\end{center}
			\caption{Performance of DS techniques and K-NN according to the hardness level of an instance considering all 30 classification datasets.}
			\label{fig:IH}
		\end{figure}

		Then, the classification accuracy of each DS technique and the K-NN are evaluated for each specific group of instances. The results of the DS techniques and K-NN according to the hardness level of the instance are presented in Figure~\ref{fig:IH}. For the sake of simplicity, we considered only the top six DS algorithms. Moreover, only the 7-NN was considered since it outperformed the 1-NN. Based on this analysis, we can see that DS methods achieve higher performance for samples with a high degree of instance hardness. When the IH level is low (IH < 0.4), the K-NN method presents the best result. However, we can see a huge drop in classification accuracy when the IH level increases. 
		
		The accuracy of the K-NN for the samples with IH = 0.7 is around 5\%, while the best DS techniques obtain an accuracy higher than 50\% for such instances (META-DES, META-DES.O and KNORA-U). Moreover, for an IH higher than 0.71 the classification accuracy of the K-NN is equal to zero, while the best DS technique obtained a much higher classification accuracy for such samples. Hence, the reasons behind the outperformance of DS techniques over the K-NN method can be explained by the fact that DS techniques can better deal with samples that are associated with a high degree of instance hardness. 
				
		We can clearly see that DS methods outperform the K-NN for the classification of samples associated with a high degree of instance hardness. This is due to the fact that a high IH value means that the majority of the samples in the neighborhood of the query instance come from a different class. Therefore, the K-NN classifier cannot predict the correct label. However, when using DS techniques, it is possible to achieve the correct prediction for such instances as long as there is at least one base classifier or a few that crosses the neighborhood of the query sample (as shown in Figure~\ref{fig:regioncompetence}). This result explains why DS techniques often outperform the K-NN classifier, even though the same neighborhood size is considered by both techniques.
		
		Hence, we are able to answer two questions posed in the paper: The reasons why DS techniques present a better performance than the K-NN is due to the fact that DS techniques can deal with samples with a high degree of instance hardness. Moreover, DS techniques should be used for the classification of instances that are associated with a high degree of instance hardness (samples that are located close to the decision border), while the K-NN should be used for the classification instances associated with a low degree of instance hardness (e.g., IH < 0.4). Moreover, for all sample associated with a high degree of IH (IH > 0.4), that were correctly classified by a DS algorithm, there was at least one base classifier in the pool crossing the region of competence (i.e., which could predict the correct label for samples of different classes).
		
		Thus, DS techniques are able to correctly classify instances that are associated with a high degree of instance hardness as long as they can select the base classifiers that are competent locally. For such a condition to be satisfied, it is required that there is at least one base classifier crossing the decision border in the local region of the query sample (as shown in Figure~\ref{fig:regioncompetence}). The classifier should also obtain a high local performance in order to be selected by the corresponding dynamic selection technique. Moreover, it would be preferable to guarantee the presence of multiple locally competent classifiers rather than just one. As the number of competent base classifiers increases, the probability of selecting only the competent ones should also increase. 
		
		\section{Conclusion}
		
		In this work, we perform an analysis comparing dynamic selection techniques with the K-NN classifier in order to better understand why and when dynamic selection techniques outperform the K-NN classifier. The analysis is motivated by the fact that the majority of the DS techniques are based on the K-NN definition, and the quality of its neighborhood has a huge impact on the performance of DS methods.
		
		Experimental results demonstrate that the majority of DS techniques obtain a significant improvement in classification performance. Moreover, an analysis conducted using instance hardness shows that the reasons in which DS presents better classification performance is due to the fact that DS techniques are better able to deal with samples with a high degree of instance hardness, while the K-NN classifier works well for samples with a low degree of instance hardness, but fails to predict the correct label for samples with a high degree of IH (the accuracy of the K-NN classifier is close to 0 for samples with an IH of 0.7). 
		
		Future work would involve the definition of a system in two steps: first the hardness of a test instance is calculated (based on its neighborhood defined over the training and validation data), and based on its hardness the system could select whether using the K-NN or applying a DS technique for classification. In this case, the DS scheme is only used to classify samples associated with a high degree of instance hardness i.e. borderline samples, while K-NN should be used for classifying samples with a low degree of instance hardness. Such approach would not only improve generalization performance, but also reduce the computational complexity involved, since the DS techniques would only be used for the classification of a few test samples.
		
		\section*{Acknowledgment}
		
		This work was supported by the Natural Sciences and Engineering Research Council of Canada (NSERC), the \'{E}cole de Technologie Sup\'{e}rieure (\'{E}TS Montr\'{e}al), CNPq (Conselho Nacional de Desenvolvimento Cient\'{i}fico e Tecnol\'{o}gico) and FACEPE (Funda\c{c}\~{a}o de Amparo \`{a} Ci\^{e}ncia e Tecnologia de Pernambuco).
		
		\bibliographystyle{IEEEtran}
		\bibliography{report}

	\end{document}